%% file: neurips_2026.tex
\documentclass{article}

\usepackage[preprint]{neurips_2026}

\usepackage[utf8]{inputenc} % allow utf-8 input
\usepackage[T1]{fontenc}    % use 8-bit T1 fonts
\usepackage{hyperref}       % hyperlinks
\usepackage{url}            % simple URL typesetting
\usepackage{booktabs}       % professional-quality tables
\usepackage{amsfonts}       % blackboard math symbols
\usepackage{nicefrac}       % compact symbols for 1/2, etc.
\usepackage{microtype}      % microtypography
\usepackage{xcolor}         % colors
\usepackage{multirow}
\usepackage[table]{xcolor}
\usepackage{graphicx}
\usepackage{booktabs}
\usepackage{amssymb}
\usepackage{amsmath}
\usepackage{float}
\usepackage{wrapfig}
\usepackage{caption}

\title{GeoPano: Towards Geometrically Accurate 3D Reconstruction from a Single Panorama}

\author{
\bfseries
Jing OU$^{1*}$ \quad
Zidong Cao$^{1*}$ \quad
Liaoyuan Fan$^3$ \quad
Zhuoxiao Li$^1$ \quad
Shuai Zhang$^1$ \\
\bfseries
Yinrui Ren$^2$ \quad
Dongli Wu$^1$ \quad
Haoang Li$^1$ \quad
Hui Xiong$^{1\dagger}$ \\
\bfseries
Wufan Zhao$^{1\dagger}$ \\
\normalfont
$^1$The Hong Kong University of Science and Technology (Guangzhou), China \\
$^2$Sun Yat-sen University, Guangzhou, China \\
$^3$The Hong Kong University, Hong Kong \\
$^*$Equal contribution \quad
$^\dagger$Corresponding author
}

\begin{document}

\maketitle

\begin{figure}[!h]
    \centering
    \includegraphics[width=1.0\linewidth]{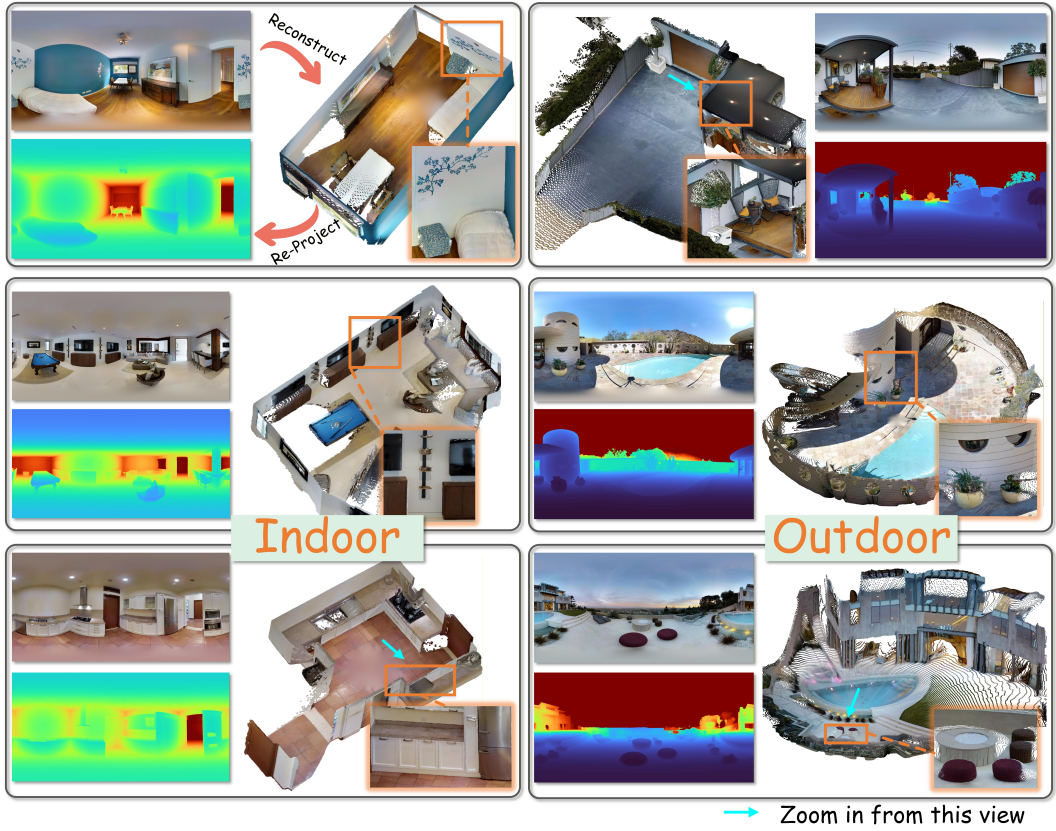}
    % \vspace{-15pt}

    \caption{For each scene, our GeoPano takes a single panorama as input, directly recovers distortion-free 3D point clouds, which can be easily re-projected into the sphere plane to obtain the ERP depth map. Given diverse in-the-wild panoramas from both indoor and outdoor scenes, GeoPano can robustly reconstruct high-fidelity 3D geometry, showing its impressive zero-shot capabilities.}
    % Given an in-the-wild panorama, \textbf{GeoPano} robustly reconstructs an \textbf{accurate} 3D point cloud and depth map. It produces \textbf{distortion-free}, \textbf{high-fidelity} geometry and generalizes seamlessly to both indoor and outdoor open-domain scenarios.}
    % \vspace{-10pt}
    \label{fig:coverfigure}
\end{figure}

\input{sec/0_abstract} 
\input{sec/1_intro} 
\input{sec/2_relatedworks}

\input{sec/3_method}

\input{sec/4_experiment}

\input{sec/5_conclusion}

\clearpage
\bibliographystyle{plainnat}
\bibliography{main}  
\clearpage
\appendix
\clearpage

\end{document}

%% file: sec/0_abstract.tex
\begin{abstract}
Panoramic 3D reconstruction enables the holistic recovery of complete 360 $^\circ$ scene geometry, offering significant benefits for robotics and AR/VR applications. 
However, directly estimating geometry from ERP images is inherently challenged by spherical distortions. To reduce spherical projection distortions, a common strategy is to decompose panoramas into distortion-free perspective views, but independently processing these views often introduces cross-view geometric inconsistencies.
To address these challenges, we introduce GeoPano, a feed-forward framework for single-view panoramic 3D reconstruction that recovers a geometrically accurate and consistent point cloud from a single panorama. Our approach is driven by two key innovations: First, we propose a panoramic-aware attention aggregator, featuring a streamlined attention architecture equipped with a novel Pano-Attention mechanism, which explicitly exploits the intrinsic spherical geometry of the panorama to guide cross-view interactions, thereby suppressing geometric inconsistencies. Furthermore, we formulate a hybrid supervision strategy to jointly enforce fine-grained local details in perspective space and global structural coherence in spherical space. Experiments demonstrate that our method significantly outperforms previous SOTA approaches in reconstruction accuracy, geometric consistency, and visual fidelity.
\end{abstract}

%% file: sec/1_intro.tex
\section{Introduction}
\label{intro}
Omnidirectional vision has garnered significant attention due to its capability to capture a comprehensive 360$^{\circ}$ field-of-view, acting as a critical enabler for holistic scene understanding \citep{ai2025survey, gao2022review}. Consequently, panoramic monocular 3D reconstruction, or depth estimation has emerged as a cornerstone task for numerous downstream applications, ranging from autonomous driving \citep{kumar2021omnidet,zheng2023both} to immersive virtual reality (VR) \citep{yang2022neural, yang2024dreamspace, cao2023omnizoomer} and visual navigation \citep{li2023panogen}. In current pipelines, the equirectangular projection (ERP) is widely used to represent spherical data on 2D planes. However, this projection is fundamentally hindered by non-uniform geometric deformations, particularly the extreme stretching concentrated at the polar regions \citep{ai2025survey}. Such inherent spherical distortions disrupt the expected spatial distribution of visual features, causing conventional perspective 3D reconstruction to degrade significantly when applied directly to ERP imagery.

To circumvent these projection artifacts, a prevailing strategy utilizes perspective projection to extract a set of distortion-free patches from the panorama. Nevertheless, processing these extracted views in isolation disrupts global contextual dependencies, invariably leading to cross-view inconsistencies and boundary seams. Existing methods attempt to resolve this fragmentation through two distinct paradigms: end-to-end supervised fusion and zero-shot training-free inference.
End-to-end supervised architectures \citep{wang2020bifuse, wang2022bifuse++, jiang2021unifuse, li2022omnifusion, ai2023hrdfuse} attempt to bridge this spatial gap by explicitly fusing global ERP context with local extracted features during training. However, their efficacy is heavily bottlenecked by the prohibitive cost of dense 360$^{\circ}$ depth annotations \citep{chang2017matterport3d, armeni2017joint, zheng2020structured3d}, resulting in limited cross-domain generalization. Furthermore, the massive computational overhead of multi-branch networks typically restricts these models to low-resolution inputs. Conversely, zero-shot training-free approaches \citep{rey2022360monodepth, peng2023high, jung2025rpg360} bypass domain-specific training by leveraging pre-trained perspective foundation models \citep{ranftl2020towards, yin2021learning} to process the decomposed perspective patches independently. While this effectively avoids the need for 360$^{\circ}$ datasets, these methods face their own critical bottlenecks: reconciling scale ambiguities and enforcing cross-view alignment heavily relies on either computationally expensive iterative post-processing \citep{jung2025rpg360, rey2022360monodepth} or auxiliary panoramic models as references \citep{peng2023high}.

Currently, the advent of feed-forward 3D reconstruction architectures (e.g., VGGT \citep{wang2025vggt} and $\pi^3$ \citep{wang2025pi}) has established a new frontier in visual geometry, operating as 3D foundation models. These networks excel at reconstructing globally consistent 3D structures from uncalibrated multi-view images by intrinsically capturing cross-view geometric relations. Motivated by this, we reformulate panoramic 3D reconstruction as a sparse multi-view problem via view decomposition, enabling us to directly leverage the robust geometric priors of foundation models for holistic scene recovery. Furthermore, benefiting from extensive pre-training on millions of real-world 3D scenes, these foundation models inherently possess exceptional sparse-view geometric reasoning capabilities. Such a rich, data-driven prior allows these frameworks to readily adapt to a sparse decomposed-view setup requiring only minimal fine-tuning data.

Fundamentally, the multi-view reasoning prowess of these foundation models is rooted in their strong attention aggregators. To adapt them for panoramic scenes, we analyze how their attention aggregator captures global information across decomposed panoramic views. This exploration exposes two bottlenecks: (1) early global attention in $\pi^3$ fails to capture geometric correlations such as spatial overlap; and (2) while later global attention layers exhibit a clear preference for interacting within overlapping views, interactions stemming from entirely non-overlapping views remain significant, which we argue is detrimental to achieving cross-view consistency.

To directly address these two bottlenecks, we propose a novel panoramic-aware Attention Aggregator. Specifically, to prevent the ineffective and diffuse interactions observed in early layers, our architecture initially employs a sequence of intra-frame self-attention layers. This stage focuses entirely on extracting reliable local geometry, explicitly bypassing premature cross-view interference. Subsequently, to tackle the residual noise in later stages, we introduce alternating attention layers augmented with our Pano-Attention. Integrated via residual connections, this Pano-Attention explicitly guides query tokens to interact strictly with geometrically overlapping regions across the decomposed views. Consequently, it effectively suppresses irrelevant interactions and significantly enhances globally consistent scene reconstruction. 
In summary, our contributions are three-fold:
\textbf{(1) A Panoramic-Aware Attention Aggregator} that explicitly incorporates spherical geometry to guide cross-view feature interactions. This design mitigates irrelevant token interactions and improves geometric consistency across decomposed perspective views. \textbf{(2) A hybrid supervision strategy} that jointly optimizes local perspective-space point clouds and global spherical-space structure. This strategy preserves fine-grained local geometry while enhancing globally consistent 3D reconstructions. \textbf{(3) }Extensive experiments on standard benchmarks demonstrate that our method achieves superior performance. It consistently outperforms prior single-view panoramic approaches in reconstruction accuracy, geometric consistency, and overall visual fidelity.

%% file: sec/2_relatedworks.tex
\section{Related Work}
\subsection{Panoramic Monocular Depth Estimation}

Panoramic monocular depth estimation reconstructs holistic 3D scene depth from single equirectangular projection (ERP) images. Current solutions generally follow two main tracks: direct ERP-domain estimation and multi-view fusion.

\noindent \textbf{ERP-domain estimation.}
ERP-based methods preserve global consistency but suffer from spherical distortions and data scarcity. To mitigate distortions, prior works employ architectural adaptations like distortion-aware convolution and spherical encoding \citep{cao2024monocular, eder2019convolutions, sun2021hohonet, tateno2018distortion, yun2023egformer}. To address data limitations, recent studies focus on data synthesis and knowledge transfer, such as generating pseudo-labels from perspective estimators \citep{cao2024panda}, expanding datasets via diffusion-based outpainting \citep{li20252}, or introducing large-scale panoramic datasets \citep{lin2025depth}. Despite these efforts, inherent distortions and data scarcity continue to hinder depth accuracy in highly distorted regions.

\noindent \textbf{Multi-view Fusion.}
Conversely, multi-view fusion paradigms mitigate spherical distortions by decomposing panoramas into perspective views \citep{rey2022360monodepth, ai2023hrdfuse, jiang2021unifuse, li2022omnifusion, mohadikar2024ms360, wang2018self, wang2020bifuse, wang2022bifuse++}. To address spatial fragmentation, recent methods explore advanced fusion strategies, such as post-processing optimization for view alignment \citep{jung2025rpg360} or representing static panoramas into sequential video frames to utilize video depth models \citep{cao2025st}. Despite these efforts, the multi-view fusion strategy inherently disrupts global structural consistency. Consequently, these approaches often rely on time-consuming post-processing or cross-domain feature fusion, which struggles to enforce accurate global consistency in complex scenes.

\subsection{Feed-forward 3D Reconstruction Models}
\label{subsec:panoramic-3d-models}
DUST3R~\citep{wang2024dust3r} introduces a transformative paradigm by directly regressing 3D point maps and camera poses from uncalibrated images. Subsequent works such as Fast3R~\citep{yang2025fast3r} and SLAM3R~\citep{liu2025slam3r} extend this to multi-view and video inputs via global fusion transformers. Scaling this further, VGGT~\citep{wang2025vggt} leverages massive 3D datasets to achieve robust in-the-wild reconstruction. To resolve the performance degradation caused by reference-frame dependency in these architectures, ${\pi}^3$~\citep{wang2025pi} proposes a permutation-equivariant design. Despite their generalization strengths, these regression-based methods often lack explicit geometric constraints tailored for the spherical topology of panoramas, potentially leading to structural drift in 360$^\circ$ reconstruction.

Recently, researchers have extended this feed-forward paradigm to the spherical domain for holistic $360^\circ$ reconstruction.
PanoSplatt3R~\citep{ren2025panosplatt3r} adapts perspective pre-training for unposed wide-baseline panorama 3D reconstruction, enabling high-fidelity novel view synthesis. PanoVGGT~\citep{guo2026panovggt} proposes an end-to-end Transformer to jointly predict camera poses, dense depth, and 3D point clouds directly from unordered panoramas by mitigating spherical distortions. VGGT-360~\citep{yuan2026vggt} achieves zero-shot panoramic depth estimation by reformulating the task as a 3D reconstruction problem utilizing perspective foundation models.

%% file: sec/3_method.tex
\section{Methodology}

\begin{figure*}
    \centering
    \includegraphics[width=1.0\linewidth]{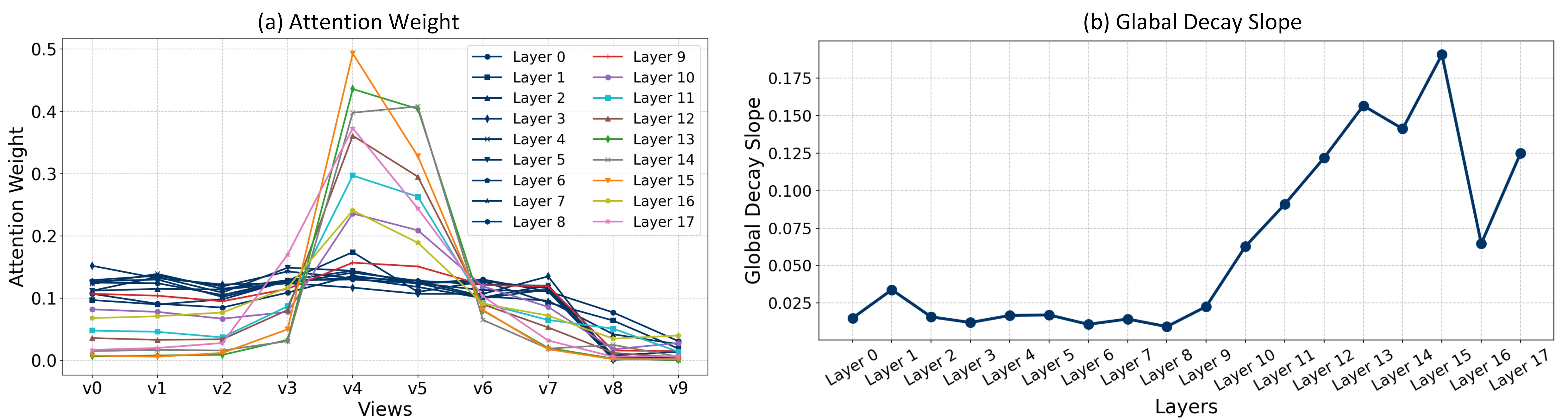}
    \caption{Statistics of global attention cross-view interactions in $\pi^3$.}
    \vspace{-10pt}
    \label{fig:attention}
\end{figure*}
\subsection{Analysis of Cross-View Interactions}
\label{subsec:analy_atten}
We adopt $\pi^3$~\citep{wang2025pi} as our base model, owing to its strong robustness in sparse-view 3D reconstruction. Given $N$ perspective views, $\pi^3$ utilizes a DINOv2 encoder~\citep{oquab2023dinov2} to extract feature tokens. These tokens are then processed by an attention aggregator comprising 18 alternating intra-frame and global attention blocks for feature aggregation. Finally, camera and point map heads decode the aggregated features into camera poses and 3D point maps.

To better adapt the multi-view geometric reasoning capability of $\pi^3$ to panoramic reconstruction, we analyze the behavior of global attention across different layers of the attention aggregator, inspired by the analysis in AVGGT~\citep{sun2025avggt}. We conduct an exploratory study on a small-scale dataset, which ultimately motivates the design of our proposed attention aggregator. Specifically, we select one token from view 4 as the query token. For each global attention layer, we separately extract the top-100 key/value tokens with the highest attention weights for this query token, and then compute their view-wise distribution. 

As illustrated in Fig.~\ref{fig:attention} (a), in the early global attention layers (Layers 0 $\sim$ 8, blue in the figure), the attention weights are uniformly distributed across the first eight horizontal views (v0 $\sim$ v7). The v8 (upward) and v9 (downward) receive minimal attention likely because these low-texture regions (e.g., skies or grounds) semantically diverge from the horizontal views. Consequently, this early-stage attention exhibits no preference for views with actual spatial overlap, resulting in a diffuse distribution.
% Perspective views decomposed from a panorama differ from conventional multi-view images in two key aspects: they share a common camera center, and their relative rotations can be substantially larger. As a result, cross-view token interactions are governed mainly by spherical geometry rather than by standard multi-view parallax. This motivates us to examine when and how global attention should be introduced in the network.

% As suggested by AVGGT~\citep{sun2025avggt}, global attention in early network layers may not contribute effectively and can even be detrimental to multi-view feature aggregation. In our configuration, perspective views decomposed from a panorama exhibit distinct geometric properties compared to conventional multi-view images. First, these decomposed views lack translational displacement, as they all share an identical camera projection center; Second, the relative rotational difference between adjacent views is significantly larger than in standard multi-view setups. 

% To verify whether the findings of AVGGT still hold under these extreme geometric constraints, 

In contrast, the later global attention layers (Layers 9 $\sim$ 17, other colors in Fig.~\ref{fig:attention} (a)) demonstrate an explicit interaction bias. The attention becomes highly concentrated on the source view (v4) and its spatially overlapping neighbors (v5), while interactions with irrelevant views are suppressed. The Global Decay Slope metric quantitatively confirms this trend. A higher slope means the attention is more tightly focused. As shown in Fig.~\ref{fig:attention} (b), the slope increases significantly in later layers, proving that the model successfully concentrates on geometrically relevant regions.

Consequently, we conclude that under decomposed perspective views, early global attention in $\pi^3$ fails to capture geometric correlations such as spatial overlap. Furthermore, we find that even in the later global attention layers, interactions stemming from entirely non-overlapping views are still significant, which we argue is detrimental to achieving cross-view consistency. Motivated by these two findings, we subsequently introduce the design of our attention aggregator, aiming to enhance the efficiency of cross-view feature interactions and improve the global consistency of the reconstruction.

\begin{figure*}
    \centering
    \includegraphics[width=0.9\linewidth]{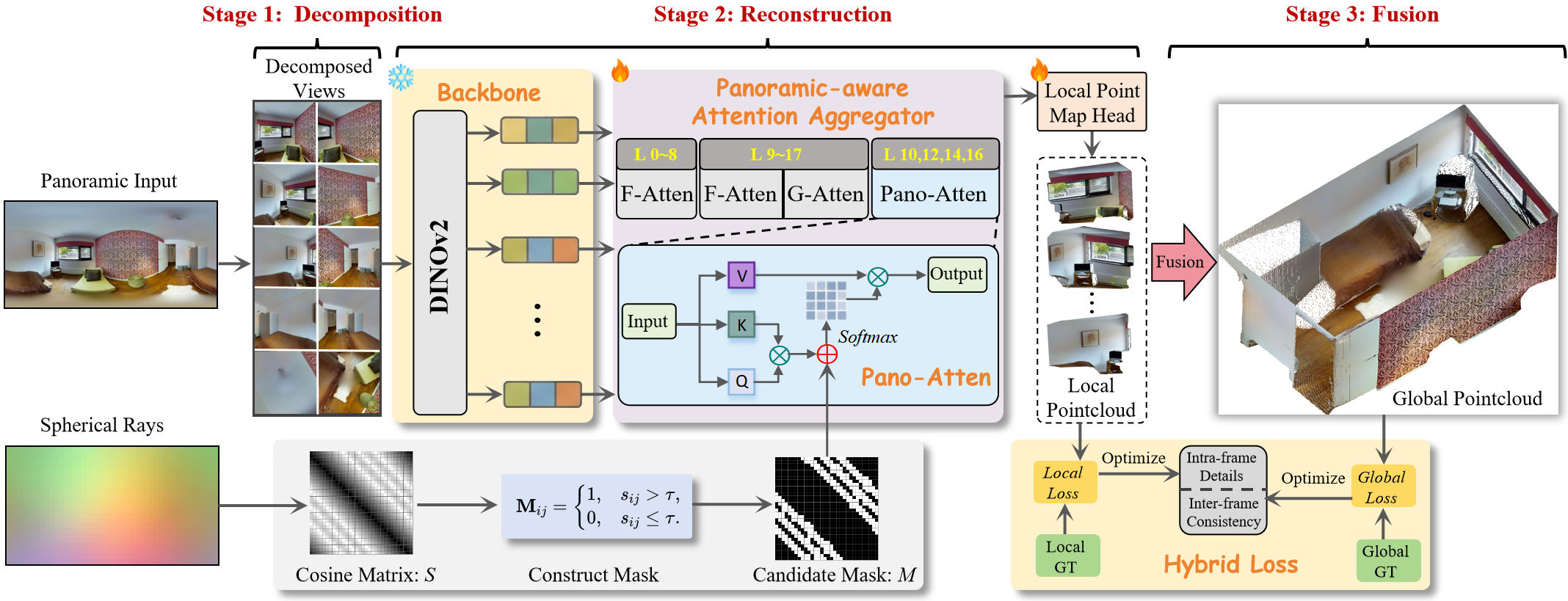}
    % \vspace{-15pt}
    \caption{Overview of the framework.
Given a panorama, we decompose it into perspective views and encode them using DINOv2. The extracted features are then aggregated by a \textbf{panoramic-aware attention aggregator}. Based on the aggregated features, the model predicts local point maps, which are fused into a global point cloud. We further introduce a \textbf{hybrid loss} to jointly enforce local details and global consistency. G-Attn and F-Attn denote Global Attention and Frame Attention.}

    \vspace{0 pt}
    \label{fig:framework}
\end{figure*}

\subsection{Framework Overview}
\begin{figure*}
    \centering
    \includegraphics[width=0.6\linewidth]{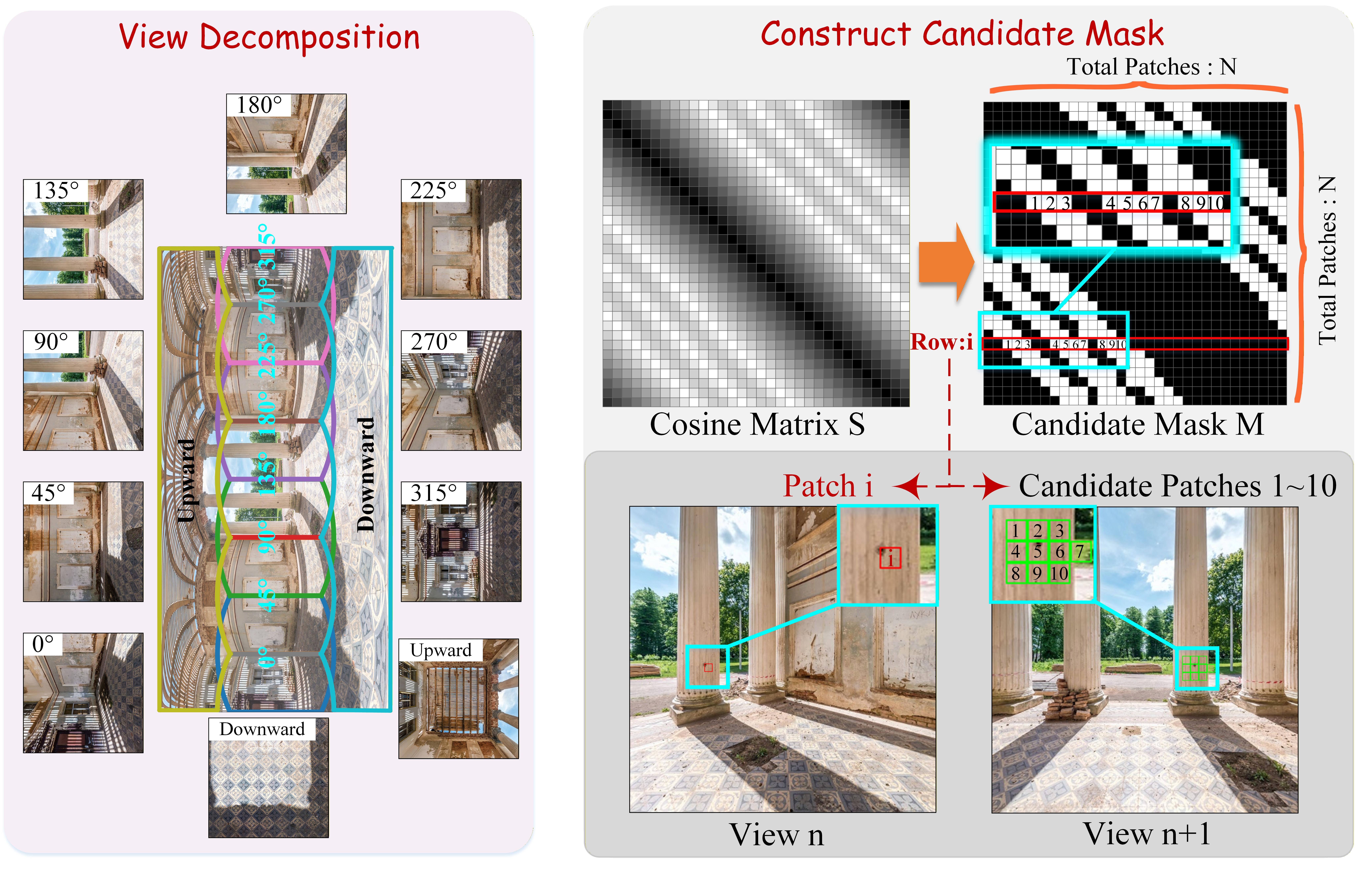}
    \caption{Left: View decomposition. The input panorama is projected into 10 perspective views: 8 evenly distributed horizontally, plus upward and downward views. Right: Token mask construction. Candidate masks are constructed by thresholding the pairwise cosine similarity of token ray directions.}
    \vspace{-10pt}
    \label{fig:creat_mask}
\end{figure*}

As illustrated in Fig.~\ref{fig:framework}, our framework operates through a three-stage pipeline: view decomposition, local 3D reconstruction, and panoramic fusion. The input panorama is first decomposed into 10 overlapping perspective views (8 horizontal views, 1 upward, and 1 downward, as detailed in Fig.~\ref{fig:creat_mask}). During the reconstruction phase, these decomposed views are tokenized into high-dimensional features by DINOv2 \citep{oquab2023dinov2}. These tokens are subsequently processed by our proposed panoramic-aware attention aggregator, which explicitly guides multi-view interactions to enforce geometric consistency. A local point map head then decodes these aggregated features into per-view 3D point clouds. Finally, these per-view point clouds are re-projected back into the spherical coordinate system and seamlessly fused into a globally aligned, holistic 3D point cloud. The entire framework is optimized via a hybrid loss, which simultaneously supervises the local perspective point clouds and the global panoramic point cloud.

\subsection{Panoramic-aware Attention Aggregator}

\noindent \textbf{Base Architecture.}
Motivated by the findings in Sec.\ref{subsec:analy_atten}, the \textbf{panoramic-aware attention aggregator} is specifically designed to address two key issues: (1) preventing ineffective and diffuse interactions in the early global attention layers, and (2) suppressing residual interactions across non-overlapping regions in the later stages. As shown in Fig.~\ref{fig:framework}, we design the base architecture with nine layers of intra-frame self-attention followed by nine layers of alternating attention. In layers 10, 12, 14, and 16 of the Attention Aggregator, we compute Pano-Attention and add it to the output of the alternating attention via a residual connection. The first nine layers focus on extracting reliable intra-frame geometry, while the subsequent alternating attention layers jointly refine inter-frame consistency and fine-grained intra-frame features. Importantly, the introduced Pano-Attention explicitly guides the query tokens to interact with tokens in cross-view overlapping regions, thereby enhancing geometry-aware feature interactions and suppressing interactions with irrelevant tokens.

\noindent \textbf{Pano-Attention.}
We introduce Pano-Attention, which restricts token interactions to geometrically coherent regions. Specifically, As shown in Fig.~\ref{fig:creat_mask} we construct a candidate token mask $\mathbf{M}$, where the $i$-th row specifies the key/value tokens (candidate tokens 1$\sim$10 in the figure) that can be queried by the $i$-th query token. We apply this mask to the attention logits before the softmax operation, as illustrate in Fig.~\ref{fig:framework}. We explicitly mask invalid token pairs by setting their pre-softmax scores to $-\infty$, ensuring their resulting attention weights are strictly zero. In this way, each token only aggregates information from geometrically valid candidates, reducing unreliable interactions and improving cross-view consistency. 

To construct the candidate mask, we leverage rays in the spherical space to introduce global geometric information. Specifically, since all decomposed perspective views share an identical camera projection center, two patches from adjacent views are considered geometrically aligned if they correspond to the same spherical ray direction. Based on this property, we can establish accurate correspondences between any two patches, determining whether they are geometrically relevant in the spherical space.

Formally, we quantify this geometric relevance by measuring the cosine similarity between their associated spherical rays. To construct the candidate mask $\mathbf{M}$, as shown in Fig.~\ref{fig:creat_mask}, we first compute the cosine similarity $s_{ij}$ between the ray of each token $i$ and token $j$ to form a global cosine similarity matrix $\mathbf{S}$. Subsequently, the candidate mask $\mathbf{M}$ is calculated via Eq.~\ref{eq:mask}:

\begin{equation}
s_{ij} =
\frac{\mathbf{r}_i^\top \mathbf{r}_j}
{\|\mathbf{r}_i\|_2 \, \|\mathbf{r}_j\|_2},
\quad
\mathbf{M}_{ij} =
\begin{cases}
1, & s_{ij} > \tau, \\
0, & s_{ij} \leq \tau,
\label{eq:mask}
\end{cases}
\end{equation}

where $\tau$ is a threshold used to determine whether these two tokens are geometrically correlated.

\subsection{Hybrid Loss}
Building upon $\pi^3$ \citep{wang2025pi}, we first apply a local point cloud loss to supervise the fine-grained geometry of each decomposed view. However, optimizing solely in isolated perspective spaces inevitably fractures global spatial alignment. To overcome this, we introduce a panoramic point cloud loss in spherical space, effectively coupling local precision with holistic consistency.
The total loss $\mathcal{L}_{\text{total}}$ is defined as a weighted combination of the local point cloud loss $\mathcal{L}_{\text{local}}$ and the panoramic point cloud loss $\mathcal{L}_{\text{pano}}$:

\begin{equation}
\mathcal{L}_{\text{total}} = \lambda \, \mathcal{L}_{\text{local}} + \beta \, \mathcal{L}_{\text{pano}},
\end{equation}

where $\lambda$ and $\beta$ control the relative contributions of the local and panoramic losses.

\noindent \textbf{Local Point Cloud Loss.} We first compute an optimal scale $s^*_{\text{local}}$ by aligning the predicted points 
$\hat{P}_{i}^{\text{local}}$ with the ground-truth points $P_{i}^{\text{local}}$ with inverse depth weights $\frac{1}{z_{i}}$. 
Based on the scale-aligned points, we compute the local point cloud loss $\mathcal{L}_{\text{local}}$:

\begin{align}
s^*_{\text{local}} = \arg\min_{s} \sum_{i=1}^{N} \frac{1}{z_{i}} \| s\hat{P}_{i}^{\text{local}} - {P}_{i}^{\text{local}} \|_1,  & \qquad 
\mathcal{L}_{\text{local}} = \sum_{i=1}^{N} \frac{1}{z_{i}} \| s^*_{\text{local}}\hat{P}_{i}^{\text{local}} - {P}_{i}^{\text{local}} \|_1
\end{align}

\noindent \textbf{Panoramic Point Cloud Loss.} 
For panoramic point cloud loss, We first project the predicted perspective-view point clouds $\hat{P}_v$ 
from multiple views back into a spherical coordinate system using the inverse projection operator $\mathcal{F}^{-1}$:

{\small
\begin{align}
\hat{P}^{\text{pan}} &= \frac{1}{V} \sum_{v=1}^{V} \mathcal{F}^{-1}(\hat{P}_v), & \! s^*_{\text{pano}} = \arg\min_{s} \sum_{i=1}^{N} \frac{1}{d_{i}} \| s\hat{P}_{i}^{\text{pano}} - {P}_{i}^{\text{pano}} \|_1,  & \:
\mathcal{L}_{\text{pano}} = \sum_{i=1}^{N} \frac{1}{d_{i}} \| s^*_{\text{pano}}\hat{P}_{i}^{\text{pano}} - {P}_{i}^{\text{pano}} \|_1
\end{align}
}
where $V$ denotes the total number of decomposed views. Subsequently, we align the predicted point cloud $\hat{P}_{i}^{\text{pano}}$ and ground-truth point cloud ${P}_{i}^{\text{pano}}$ within the spherical space, weighted by the panoramic inverse depth $\frac{1}{d_{i}}$, to derive the optimal scale factor $s^*_{\text{pano}}$. Based on these scale-aligned points, we then compute the panoramic point cloud loss $\mathcal{L}_{\text{pano}}$.
% As illustrated in Fig.~\ref{fig:xxx}, this inverse projection mechanism explicitly propagates the spatial uniformity inherent in perspective views to the global panoramic space. Consequently, it effectively avoids the supervision imbalance caused by severe spherical distortion in the Equirectangular Projection (ERP) format.

% 这里加一个图表示，一个token可以显示的与他的对应token交互

%% file: sec/4_experiment.tex
\section{Experiments} \label{sec:exp}
\subsection{Experiment Setup} \label{sec:exp_set}
\noindent \textbf{Training Datasets.}
For training, we use Holo360D \citep{ou2026holo360dlargescalerealworlddataset}, a large-scale real-world panoramic 3D dataset comprising 75 indoor and outdoor scenes. It provides panoramic RGB images with corresponding depth maps, each captured at a resolution of 2880 × 1440. We train our model using 23,000 panoramic frames.

\noindent \textbf{Evaluation Datasets and Evaluation Metrics.} 
For evaluation, following prior work \citep{lin2025depth}, we test our model on three 360$^\circ$ depth datasets: Matterport3D \citep{chang2017matterport3d} (indoor) and Stanford2D3D \citep{armeni2017joint} (indoor), and Deep360 \citep{li2022mode} (outdoor). 
To comprehensively assess model performance, we evaluate both point cloud using 3D metrics \citep{jung2025rpg360} such as Chamfer Distance, F-Score, and Intersection over Union (IoU), and depth using standard 2D metrics including Absolute Relative Error (AbsRel), Root Mean Squared Error (RMSE), and the accuracy under a threshold $\delta_i$ with $i$=1.25.

\noindent \textbf{Implementation details.} 
For training, each model is trained using four NVIDIA A800 GPUs for 20 epochs, with each epoch consisting of 1,000 iterations. During evaluation, inference is performed on a single A800 GPU, while monitoring GPU memory usage and runtime performance.

\subsection{Qualitative and Quantitative Evaluation} \label{sec:eval}
\noindent \textbf{Comparative Baselines.}
We compare GeoPano with recent state-of-the-art panoramic approaches. We consider both methods operating directly on the equirectangular projection (ERP) format (PanDA \citep{cao2024panda}, DA$^2$ \citep{li20252}, DAP \citep{lin2025depth},  and PanoVGGT \citep{guo2026panovggt}) and those relying on view decomposition into pinhole projections (360MonoDepth \citep{rey2022360monodepth}, RPG360 \citep{jung2025rpg360}). PanoVGGT is evaluated only on Deep360, as its custom dataset splits for Matterport3D and Stanford2D3D are not publicly available, preventing a fair comparison on these benchmarks.

\begin{table*}[!ht]
\centering
\small
\setlength{\tabcolsep}{4pt}
\renewcommand{\arraystretch}{1.15}
\caption{Quantitative evaluation on 3D and 2D metrics. ERP and Pinhole denote methods using ERP projection images and pinhole images decomposed from panoramas as input, respectively.}
\label{tab:compari}

\resizebox{\textwidth}{!}{
\begin{tabular}{lcccc|ccc |ccc|ccc |ccc|ccc}
\toprule
\multicolumn{2}{c}{\textbf{Method}}
& \multicolumn{6}{c}{\textbf{Matterport3D (Indoor)}}
& \multicolumn{6}{c}{\textbf{Stanford2D3D (Indoor)}}
& \multicolumn{6}{c}{\textbf{Deep360 (Outdoor)}} \\
\cmidrule(lr){1-2}\cmidrule(lr){3-8}\cmidrule(lr){9-14}\cmidrule(lr){15-20}
& & \multicolumn{3}{c|}{\textbf{3D Metrics}} & \multicolumn{3}{c}{\textbf{2D Metrics}}
& \multicolumn{3}{c|}{\textbf{3D Metrics}} & \multicolumn{3}{c}{\textbf{2D Metrics}}
& \multicolumn{3}{c|}{\textbf{3D Metrics}} & \multicolumn{3}{c}{\textbf{2D Metrics}} \\
\cmidrule(lr){3-5}\cmidrule(lr){6-8}\cmidrule(lr){9-11}\cmidrule(lr){12-14}\cmidrule(lr){15-17}\cmidrule(lr){18-20}
\textbf{Model} & \textbf{Input}
& \textbf{Chamfer}$\downarrow$ & \textbf{F-Score}$\uparrow$ & \textbf{IoU}$\uparrow$ &AbsRel$\downarrow$ &RMSE$\downarrow$ &$\delta_1$ $\uparrow$
& \textbf{Chamfer}$\downarrow$ & \textbf{F-Score}$\uparrow$ & \textbf{IoU}$\uparrow$ &AbsRel$\downarrow$ &RMSE$\downarrow$ &$\delta_1$ $\uparrow$
& \textbf{Chamfer}$\downarrow$ & \textbf{F-Score}$\uparrow$ & \textbf{IoU}$\uparrow$ &AbsRel$\downarrow$ &RMSE$\downarrow$ &$\delta_1$ $\uparrow$\\
\midrule

PanDA & \multirow{3}{*}{ERP}
& 0.553 & 26.270 & 15.710 &0.0888 &0.3325 &0.9209
& 0.727 & 15.990 & 8.850 &0.1648 &0.2364 &0.7326
& 2.350 & 18.420 & 10.810 &0.3319 & 10.7668& 0.4498\\

DA$^2$ &
& \cellcolor[HTML]{FFCE93}0.180 & \cellcolor[HTML]{FFCE93}77.450 & \cellcolor[HTML]{FFCE93}67.700 &\cellcolor[HTML]{FFCE93}0.0667 &\cellcolor[HTML]{FFCE93}0.2882 &\cellcolor[HTML]{FFCE93}0.9561 
& \cellcolor[HTML]{FFCE93}0.145 & \cellcolor[HTML]{FFCE93}80.310 & \cellcolor[HTML]{FFCE93}69.360 &\cellcolor[HTML]{FFCE93}0.0723 &\cellcolor[HTML]{FFCE93}0.1400 &\cellcolor[HTML]{FFCCC9}0\textbf{.9545}
& 4.036 & 3.520 & 1.920 &0.1991 & 12.3898 & 0.8155\\

DAP &
& 0.269 & 63.260 & 49.930 &0.1186 &0.7510 &0.8518
& 0.487 & 33.120 & 20.600 &0.0921 &0.3820 &0.9135
& 2.214 & 34.180 & 21.850 & \cellcolor[HTML]{FFCE93}0.1213 &7.2308 &\cellcolor[HTML]{FFCCC9}\textbf{0.9159} \\

PanoVGGT &
& - & - & - &- &- &-
& - & - & - &- &- &-
&\cellcolor[HTML]{FFCE93} 1.537 &\cellcolor[HTML]{FFCE93} 69.094 &\cellcolor[HTML]{FFCE93} 55.587 & 0.1618  &\cellcolor[HTML]{FFCE93} 6.1886 & 0.7561\\
\midrule

360MonoDepth & \multirow{3}{*}{Pinhole}
& 0.632 & 28.050 & 16.990 &0.2083 &0.7909 &0.6558
& 0.576 & 25.230 & 14.670 &0.1650 &0.2823 &0.7456
& 3.761 & 25.450 & 15.080 &1.5836 & 17.2300 & 0.1412\\

RPG360 &
& 0.337 & 58.810 & 44.910 &0.2030 & 0.6670 & 0.8590
& 0.206 & 73.910 & 61.880 &0.1016&0.1924&0.9104
& 3.430 & 62.990 & 48.110 &0.2375 & 10.0562 & 0.6377\\

% $\pi^3$ &
% & 0.275 & 58.67 & 44.42 &0.2030 & 0.6670 & 0.8590
% & 0.267 & 58.64 & 43.41 &0.1016&0.1924&0.9104
% & 3.430 & \cellcolor[HTML]{FFCE93}62.990 & \cellcolor[HTML]{FFCE93}48.110 &0.2375 & 10.0562 & 0.6377\\

\textbf{Ours} &
& \cellcolor[HTML]{FFCCC9}\textbf{0.153} & \cellcolor[HTML]{FFCCC9}\textbf{82.250} & \cellcolor[HTML]{FFCCC9}\textbf{72.920} &\cellcolor[HTML]{FFCCC9}\textbf{0.0597}&\cellcolor[HTML]{FFCCC9}\textbf{0.2695}&\cellcolor[HTML]{FFCCC9}\textbf{0.9648}
& \cellcolor[HTML]{FFCCC9}\textbf{0.127} & \cellcolor[HTML]{FFCCC9}\textbf{85.040} & \cellcolor[HTML]{FFCCC9}\textbf{75.670} &\cellcolor[HTML]{FFCCC9}\textbf{0.06361}& \cellcolor[HTML]{FFCCC9}\textbf{0.1326}&\cellcolor[HTML]{FFCE93}0.9539
& \cellcolor[HTML]{FFCCC9}\textbf{1.202} & \cellcolor[HTML]{FFCCC9}\textbf{80.560} & \cellcolor[HTML]{FFCCC9}\textbf{68.710} &\cellcolor[HTML]{FFCCC9}\textbf{0.1130}&\cellcolor[HTML]{FFCCC9}\textbf{5.5837}&\cellcolor[HTML]{FFCE93}0.8289\\
\bottomrule
\end{tabular}
}
\end{table*}

\begin{figure*}[!ht]
    \centering
    \includegraphics[width=0.95\linewidth]{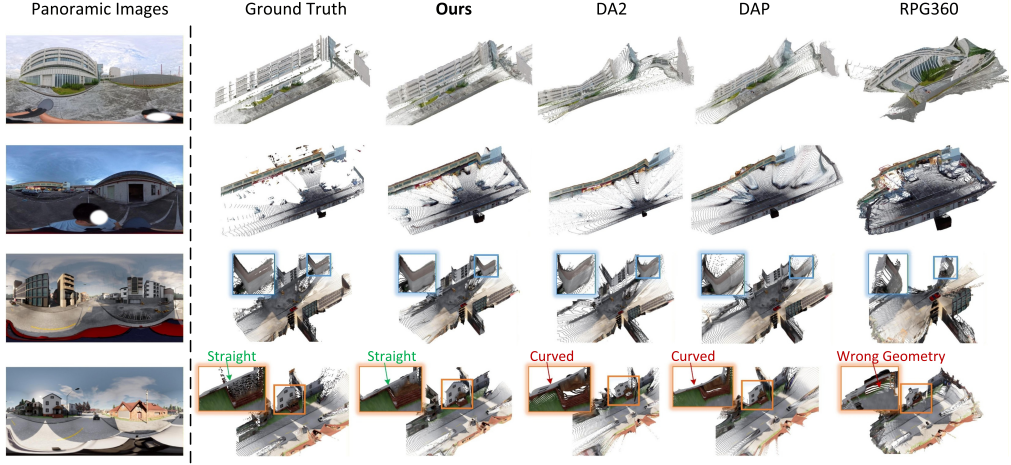}
    \caption{Qualitative comparison in \textbf{outdoor} scenarios. We compare GeoPano against DA$^2$, DAP, and RPG360. As illustrated, our model produces more precise 3D geometry and significantly mitigates structural distortions in long-range regions.}
    \label{fig:outdoor_recon_results}
\end{figure*}

\begin{figure*}[!ht]
    \centering
    \includegraphics[width=0.95\linewidth]{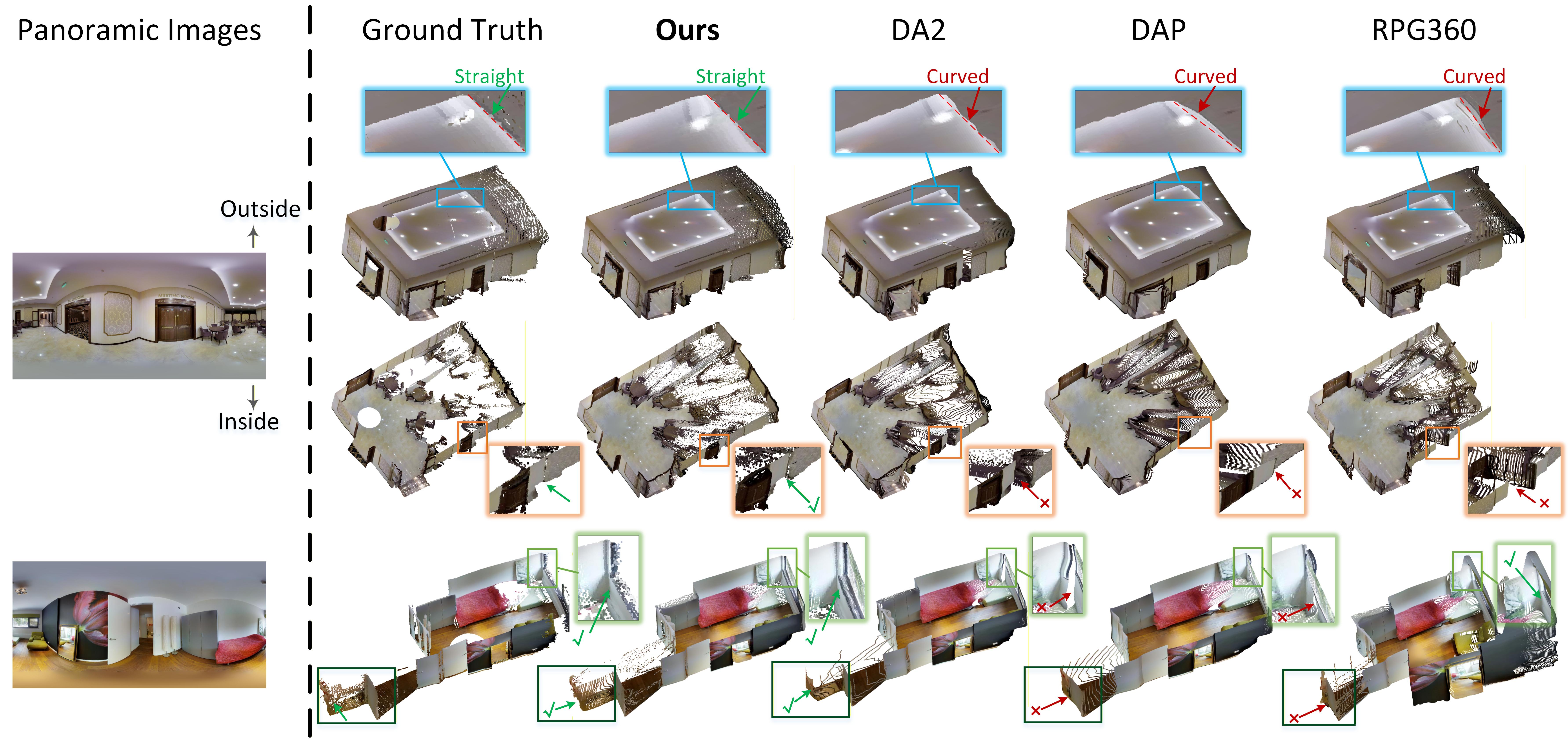}
    \caption{Qualitative comparison in \textbf{indoor} scenes. Compared to DA$^2$, DAP, and RPG360, our method achieves superior global structural coherence and recovers fine-grained local geometries.}
    \label{fig:indoor_recon_results}
\end{figure*}

\noindent \textbf{Evaluation.}
As reported in Tab.~\ref{tab:compari}, GeoPano establishes a new state-of-the-art, demonstrating superior performance across all 3D and 2D metrics. Additionally, as illustrated in Fig.~\ref{fig:outdoor_recon_results} and Fig.~\ref{fig:indoor_recon_results}, our approach significantly mitigates spherical distortions and preserves geometric consistency, yielding high-fidelity 3D structures across diverse indoor and outdoor environments.
As shown in the first two rows of Fig.~\ref{fig:outdoor_recon_results}, GeoPano achieves accurate geometric reconstructions even in challenging outdoor long-range scenes. 
% \textcolor{red}{This superiority demonstrates that effectively harnessing the strong priors from pre-trained perspective models drastically improves the framework's robustness and open-domain generalization. Notably, while RPG360 also builds upon a perspective foundation model, its heavy reliance on heuristic post-processing for spatial alignment inevitably leads to geometric degradation and structural artifacts in such demanding environments.}

%是否要提对任意位姿的鲁棒性

\subsection{Ablation Studies} \label{sec:abl}
\begin{wrapfigure}[23]{r}{0.4\textwidth}
    \centering
    \includegraphics[width=\linewidth]{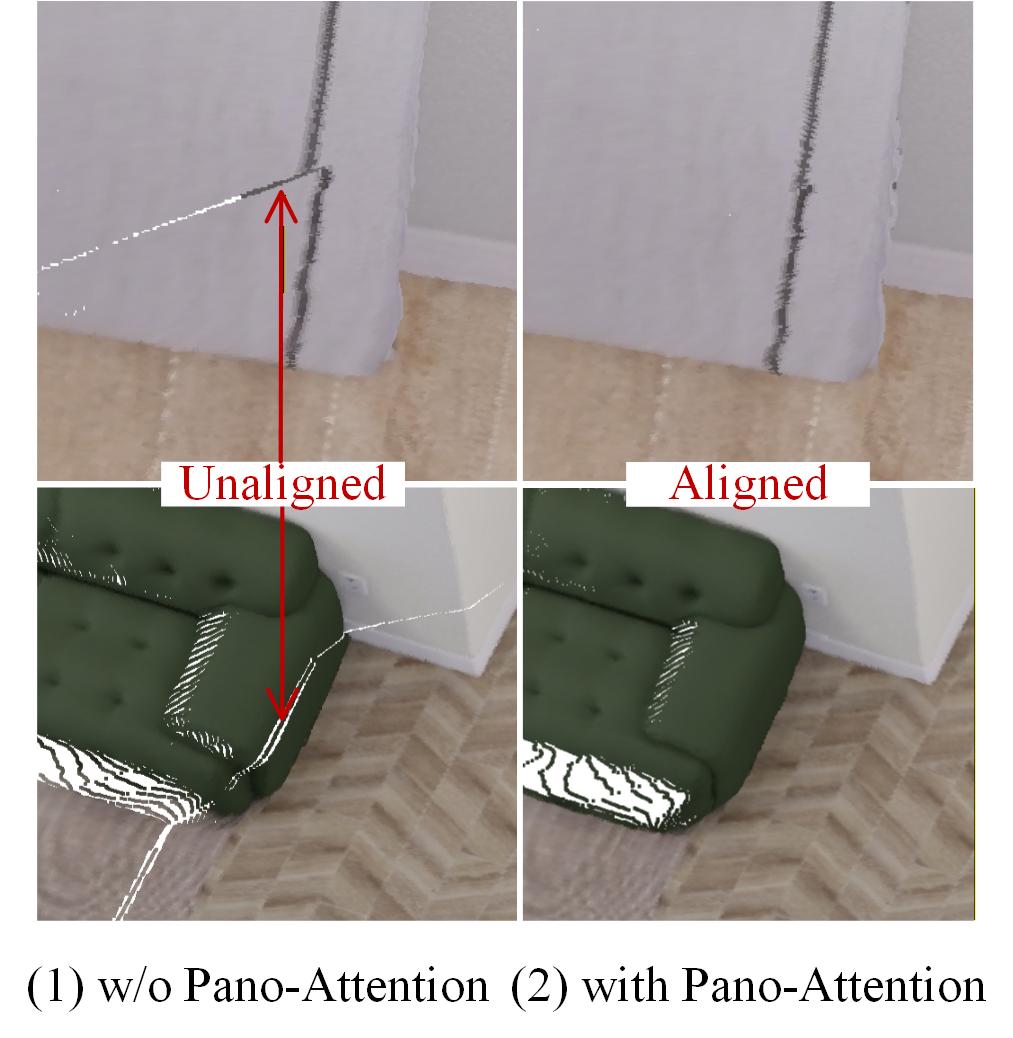}
    % \vspace{-15pt}
    \caption{Qualitative comparison with and without Pano-Attention. Integrating our proposed Pano-Attention significantly enhances the geometric consistency.}
    \vspace{-10pt}
    \label{fig:geo-attention_abl}
\end{wrapfigure}

\noindent \textbf{Main Ablation.}
To evaluate the contribution of each proposed component of GeoPano, we conduct a ablation study conducted on Matterport3D, summarized in Tab.~\ref{tab:ablation_main}. We start from a panoramic
fine-tuing version of $\pi^3$ on our training data as the baseline and add: (1) dropping early global attention, Pano-Attention, and (3) hybrid loss. All introduced modules effectively improve reconstruction accuracy; notably, Pano-Attention intrinsically maintains spatial consistency across the decomposed views, with qualitative comparisons shown in Fig.~\ref{fig:geo-attention_abl}.
\begin{table*}[ht]
    \centering
    \small
    \setlength{\tabcolsep}{4pt}
    \renewcommand{\arraystretch}{1.15}
    \caption{Ablation study of the proposed components. Each designed module effectively improves the reconstruction accuracy. Compared with the baseline, our full model reduces the inference time by 10.3\%, with only a 5.6\% increase in memory consumption.}
    \label{tab:ablation_main}

    \resizebox{1.0\textwidth}{!}{
        \begin{tabular}{ccc ccc cc}
            \toprule
            Drop Global Attention 
                & Pano-Attention 
                & Hybrid Loss 
                & Chamfer$\downarrow$ 
                & F-Score$\uparrow$ 
                & IoU$\uparrow$
                & Inference Time (s)
                & GPU Mem. (GB) \\
            \midrule
            $\times$
                & $\times$
                & $\times$
                & 0.164
                & 81.110
                & 71.640
                & 0.500
                & 4.890 \\
            $\checkmark$
                & $\times$
                & $\times$
                & 0.157
                & 81.780
                & 72.370
                & 0.370
                & 4.250 \\
            $\checkmark$
                & $\checkmark$
                & $\times$
                & 0.155
                & 82.010
                & 72.710
                & 0.450
                & 5.170 \\
            $\checkmark$
                & $\checkmark$
                & $\checkmark$
                & 0.153
                & 82.250
                & 72.920
                & 0.450
                & 5.170 \\
            \bottomrule
        \end{tabular}
    }
\end{table*}

\noindent \textbf{Effectiveness of Early Global Attention.}
We investigate the effect of dropping early global attention in the baseline $\pi^3$ architecture for single-frame panoramic 3D reconstruction, as shown in Tab.~\ref{tab:rga_abl}. Specifically, we train and evaluate several variants where global attention is dropped from the initial layers, including layers $0\sim4$, $0\sim6$, $0\sim8$, and $0\sim10$. Dropping global attention from layers $0\sim8$ yields the best trade-off between computational efficiency and geometric accuracy. This observation supports our analysis of $\pi^3$, suggesting that the diffuse interactions induced by early global attention can adversely affect global consistency in the panoramic setting. 
% Based on this result, we use only intra-view self-attention in the first $0\sim8$ layers and adopt this configuration as the base architecture of our Panoramic-Aware Attention Aggregator.

\begin{table*}[!t]
\centering
\small
\setlength{\tabcolsep}{4pt}
\renewcommand{\arraystretch}{1.15}
\caption{Ablation study on the effect of dropping early global attention layers in alternating attention.}
\label{tab:rga_abl}

\resizebox{0.8\textwidth}{!}{
\begin{tabular}{cccccc}
\toprule
Layers w/o Global Attntion & Chamfer$\downarrow$ & F-Score$\uparrow$ & IoU$\uparrow$ & Inference Time (s) & GPU Mem. (GB) \\
\midrule
Baseline  & 0.164 & 81.110 & 71.640 & 0.500 & 4.890 \\
$0\sim4$  & 0.161 & 81.530 & 71.970 & 0.440 & 4.540 \\
$0\sim6$  & \cellcolor[HTML]{FFCCC9}0.157 & \cellcolor[HTML]{FFCCC9}82.050 & \cellcolor[HTML]{FFCCC9}72.680 & 0.410 & 4.440 \\
$0\sim8$  & \cellcolor[HTML]{FFCCC9}0.157 & \cellcolor[HTML]{FFCE93}81.780 & \cellcolor[HTML]{FFCE93}72.370 & 0.390 & 4.250 \\
$0\sim10$ & 0.159 & 81.450 & 72.030 & 0.350 & 4.000 \\
\bottomrule
\end{tabular}
}
\vspace{-5 pt}
\end{table*}

\noindent \textbf{Effectiveness of Pano-Attention.}
We investigate the effectiveness of Pano-Attention and its optimal configuration. Specifically, we compare four settings: (1) no Pano-Attention, (2) Pano-Attention applied at every other layer (layers 10, 12, 14, 16), (3) Pano-Attention applied at every layer (layers 9 $\sim$ 16), and (4) standard global attention applied at every other layer (layers 10, 12, 14, 16). The results show that applying Pano-Attention at every other layer achieves the best performance, indicating that moderate geometric constraints better balance reliable cross-view interaction and global information aggregation. In contrast, the standard global attention baseline (configure 4) fails to yield comparable gains, demonstrating that the improvement stems from our geometry-constrained cross-view interaction rather than merely stacking additional attention layers.

% \noindent Building upon the previous configuration, we further compare alternative methods for injecting global information, including using ray-rotation positional encoding (RoRE), traditional ray encoding, and replacing Pano-Attention with standard global attention. Our results show that Pano-Attention consistently achieves the best reconstruction performance.

\noindent \textbf{Effectiveness of hybrid loss.}
Tab.~\ref{tab:loss_abl} validates the necessity of our dual-constraint design, demonstrating that the combination of local and global losses is strictly superior to single-loss baselines. We observe that peak performance is achieved with $\lambda=1$ and $\beta=0.75$. This optimal weighting strategy explicitly navigates the trade-off between local and global representation learning, ensuring that high-fidelity local geometric structures are flawlessly preserved while still satisfying strict global coherence.

\begin{table*}[ht]
\centering
\small

\begin{minipage}[t]{0.65\textwidth}
    \vspace{0pt}
    \centering
    \setlength{\tabcolsep}{4pt}
    \renewcommand{\arraystretch}{1.15}
    \captionof{table}{Ablation study on the configuration of Pano-Attention layers.}
    \label{tab:geoatten_abl}

    \resizebox{\linewidth}{!}{
        \begin{tabular}{lccc}
            \toprule
            Pano-Attention Layers & Chamfer$\downarrow$ & F-Score$\uparrow$ & IoU$\uparrow$ \\
            \midrule
            (1) 0 Pano-Attention 
                & 0.157 
                & 81.780 
                & 72.370 \\
            (2) 4 Pano-Attention (10,12,14,16) 
                & \cellcolor[HTML]{FFCCC9}0.153 
                & \cellcolor[HTML]{FFCCC9}82.250 
                & \cellcolor[HTML]{FFCCC9}72.920 \\
            (3) 8 Pano-Attention (9,10,11,12,13,14,15,16) 
                & 0.156 
                & 81.590 
                & 72.190 \\
            \midrule
            (4) 4 Global Attention (10,12,14,16) 
                & 0.160 
                & 81.470 
                & 71.920 \\
            \bottomrule
        \end{tabular}
    }
\end{minipage}
\hfill
\begin{minipage}[t]{0.32\textwidth}
    \vspace{0pt}
    \centering
    \setlength{\tabcolsep}{4pt}
    \renewcommand{\arraystretch}{1.15}
    \captionof{table}{Ablation study with different loss weightings. }
    % $\lambda$ and $\beta$ denote the weights of the local and global loss, respectively.
    \label{tab:loss_abl}

    \resizebox{\linewidth}{!}{
        \begin{tabular}{lcccc}
            \toprule
            $\lambda$ & $\beta$ & Chamfer$\downarrow$ & F-Score$\uparrow$ & IoU$\uparrow$ \\
            \midrule
            1   & 0    & 0.155 & 82.010 & 72.710 \\
            1   & 0.5  & 0.154 & 82.030 & 72.670 \\
            1   & 0.75 & \cellcolor[HTML]{FFCCC9}0.153 & \cellcolor[HTML]{FFCCC9}82.250 & \cellcolor[HTML]{FFCCC9}72.920 \\
            1   & 1    & 0.156 & 81.500 & 72.000 \\
            0   & 1    & 0.154 & 81.960 & 72.550 \\
            \bottomrule
        \end{tabular}
    }
\end{minipage}

\end{table*}

%% file: sec/5_conclusion.tex
\section{Conclusion}
We proposed GeoPano, a feed-forward framework for single-view panoramic 3D reconstruction, featuring a Panoramic-Aware Attention Aggregator that leverages spherical geometry to improve cross-view consistency. A hybrid supervision strategy combines local and global point cloud losses, ensuring fine-grained details and holistic structural coherence. Experiments show that GeoPano outperforms prior single-view panoramic methods in accuracy, consistency, and visual fidelity while maintaining practical efficiency.